%% file: root.tex
\documentclass[letterpaper, 10 pt, conference]{ieeeconf}  

\IEEEoverridecommandlockouts 

\usepackage{graphicx} 
\usepackage{color}
\usepackage{soul}
\usepackage{amsmath,amssymb}
\usepackage{times}
\usepackage[table,xcdraw]{xcolor}
\usepackage{booktabs}
\usepackage{multirow}
\usepackage{cuted}
\usepackage{caption}
\usepackage[linesnumbered]{algorithm2e}
\usepackage{subcaption}
\usepackage{dsfont}
\usepackage{pgfplots}
\usepgfplotslibrary{groupplots}
\pgfplotsset{compat=1.18}

\newtheorem{observation}{Observation}
\makeatletter\let\NAT@parse\undefined\makeatother 
\usepackage[numbers,sort&compress]{natbib}
\usepackage[bookmarks=true,hidelinks]{hyperref}

\RestyleAlgo{ruled}
\SetAlgoNlRelativeSize{0}

\SetKwFunction{InitValidatedGraph}{InitializeValidatedGraph}
\SetKwFunction{InitOptimisticGraph}{InitializeOptimisticGraph}
\SetKwFunction{InitGoalQueryStates}{InitializeGoalQueryStates}
\SetKwFunction{ShortestPath}{ShortestPath}
\SetKwFunction{PlanningBudgetRemains}{PlanningBudgetRemains}
\SetKwFunction{LowestCostProposal}{LowestCostOptimisticProposal}
\SetKwFunction{PathTo}{PathTo}
\SetKwFunction{ChooseSkill}{ChooseSkill}
\SetKwFunction{BatchSampleParameters}{BatchSampleParameters}
\SetKwFunction{BatchPredict}{BatchPredict}
\SetKwFunction{MaxTransStd}{MaxPerObjectUncertainty}
\SetKwFunction{BatchInsert}{BatchInsert}
\SetKwFunction{MarkTerminalStates}{MarkTerminalStatesSatisfying}
\SetKwFunction{FirstUnvalidatedProposal}{FirstUnvalidatedProposal}
\SetKwFunction{ValidateProposal}{ValidateProposal}

\SetKwFunction{RealizedPredState}{RealizedPredecessorState}
\SetKwFunction{Reconstruct}{ReconstructFromRealizedPredecessor}
\SetKwFunction{PhysicsSimulate}{PhysicsSimulate}
\SetKwFunction{UpdatePhysicsOutcome}{UpdateFromPhysicsOutcome}
\SetKwFunction{Invalidate}{Invalidate}
\SetKwFunction{InvalidateOutgoingDependents}{InvalidateOutgoingDependents}
\SetKwFunction{IncomingConnectors}{IncomingConnectors}
\SetKwFunction{MakePartial}{MakePartial}
\SetKwFunction{RemoveProposal}{Remove}
\SetKwFunction{InsertCertified}{Insert}
\SetKwFunction{TargetGenerator}{TargetGenerator}
\SetKwFunction{CertifyConnectorInto}{CertifyConnectorInto}
\SetKwFunction{RemoveConnectorInto}{RemoveConnectorInto}
\SetKwFunction{DirectDependents}{DirectOptimisticDependents}
\SetKwFunction{Match}{Match}

\SetKw{Continue}{continue}
\SetKw{Break}{break}

\newcommand{\mosaic}{\textsc{Mosaic}}
\newcommand{\lmosaic}{\textsc{Lazy-Mosaic}}
\newcommand{\ve}[1]{\boldsymbol{#1}}
\newcommand{\Tau}{\mathrm{T}}
\newcommand{\gcert}{G_{\mathrm{cert}}}
\newcommand{\gopt}{G_{\mathrm{opt}}}
\definecolor{placeholder}{RGB}{200,90,0}

\title{\LARGE \bf \lmosaic{}: Accelerating Skill-Centric Manipulation Planning with Learned Skill-Effect World Models}
\author{
  Mukund Krishnakumar\thanks{$^{*}$ Equal contribution.}\thanks{$^{1}$ Robotics Institute, School of Computer Science, Carnegie Mellon University, Pittsburgh, PA, United States. \texttt{\{mukundkr, imishani, maxim\}@cs.cmu.edu}}$^{*,1}$,
  Itamar Mishani$^{*,1}$,
  Oren Salzman\thanks{$^{2}$ Department of Computer Science, Technion, Israel Institute of Technology, Haifa, Israel. \texttt{osalzman@cs.technion.ac.il}}$^{2}$,
  and Maxim Likhachev$^{1}$
}

\begin{document}
\bstctlcite{IEEEexample:BSTcontrol}

\maketitle
\begin{strip}       
    \centering
    \vspace{-40pt}
    \begin{minipage}{\linewidth}
        \centering
        \includegraphics[width=0.95\textwidth]{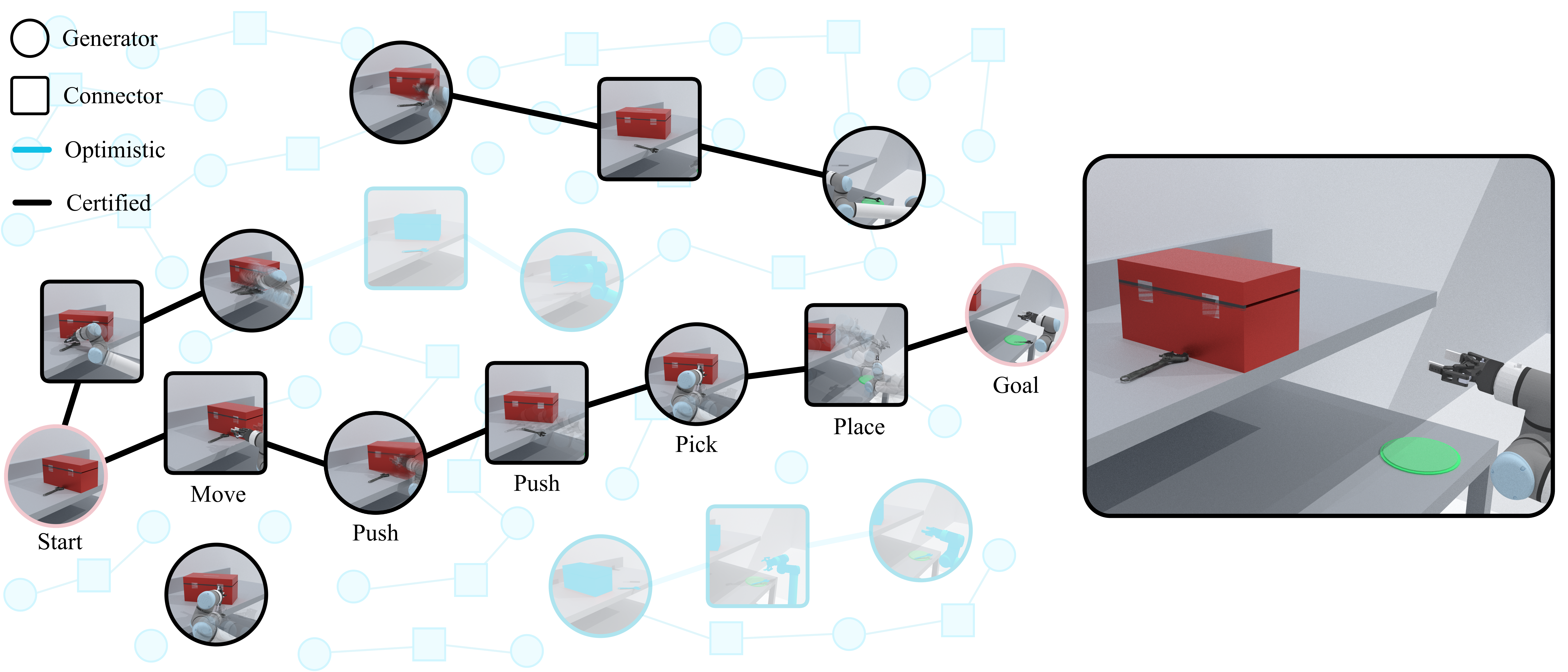}
    \end{minipage}
    
    \vspace{2.5pt}
    
    \begin{minipage}{\linewidth}
        \centering
    \end{minipage}
    \captionof{figure}{
    \textbf{Left:} \lmosaic{} performs lazy search for a sequence of parameterized skills leading from the start state to the goal (both pink), over a combined graph of certified skills and hypothesized skill outcomes. A cheap learned skill-effect world model expands the optimistic graph (faded cyan); promising hypotheses (magnified cyan) are certified in physics simulation, with successful rollouts entering the certified graph (black) and forming the complete plan. 
    The returned plans contain only certified skills. \textbf{Right:} Wrench retrieval in a semi-structured workshop. The robot must clear clutter, reposition the wrench, grasp it, and place it in the goal region (green).
    }
    \label{fig:optimistic_certified_graphs} 
\end{strip}
\thispagestyle{empty}
\pagestyle{empty}

\input{sections/abstract}

\input{sections/intro}

\input{sections/related}

\input{sections/background}

\input{sections/method}

\input{sections/experiments}

\input{sections/conclusion}

\footnotesize
\bibliographystyle{IEEEtranN}
\bibliography{root}

\end{document}

%% file: sections/abstract.tex
\begin{abstract}

Long-horizon manipulation tasks require a robot to perform a sequence of physical interactions with its environment, typically combining several different skills. Planning such behaviors efficiently is challenging: given a library of parameterized skills, solutions lie in a vast hybrid space of skill sequences and their parameters, and evaluating each candidate skill's feasibility and effect with a high-fidelity world model such as a physics simulator is computationally expensive. \mosaic{}, a recent skill-centric approach, lets the skills guide a non-sequential search toward the regions of this space where they are likely to succeed; however, it pays the full cost of simulation for every parameterized skill it considers. In this work, we introduce \lmosaic{}, a lazy planning framework that instead searches over hypothesized skill outcomes. A cheap learned skill-effect world model predicts, in a fraction of a millisecond, a skill's feasibility and the world state it produces. \lmosaic{} then uses these predictions to focus expensive simulation on promising candidate skills and, when a hypothesis is invalidated, repairs only the affected part of the search, while ensuring that every skill in a returned plan is validated in simulation. We evaluate \lmosaic{} on cluttered-shelf retrieval tasks of increasing difficulty and demonstrate it on a physical robot, showing that it plans substantially faster than baselines with fewer simulator queries, while matching or exceeding their success rate in every setting. Visit \href{skill-mosaic.github.io}{\texttt{skill-mosaic.github.io}}.

\end{abstract}

%% file: sections/intro.tex
\section{Introduction}
\label{sec:introduction}

Consider a robot tasked with retrieving a wrench from behind clutter in an automotive workshop (Fig.~\ref{fig:optimistic_certified_graphs}). This is an instance of the semi-structured settings we target in this work: the robot knows the workspace and the object types it may encounter, but which objects are present and where they rest varies from task to task. The wrench may lie beyond the gripper's reach, requiring the robot to push other objects aside and reposition the wrench before attempting a grasp. Solving such tasks with generic skills such as pushing, pulling, grasping, and transporting requires reasoning not only about which skills to compose but also about the world states they produce, since a skill that makes little immediate progress toward the goal may enable a later skill to succeed. A planner therefore needs a world model---any function that predicts the outcome of executing a parameterized skill from a given state---to compare alternative skill sequences, and such models differ in cost and fidelity. Recent skill-based planners use a physics simulator as their world model, relatively accurate but expensive to query \cite{mosaic, saleem2020selective}.

\mosaic{} \cite{mosaic} is a recently introduced planner of this kind. Rather than growing a search forward from the start state, it conducts a non-sequential search in which skills guide exploration toward regions of the state space where they are likely to succeed. \emph{Generator} skills, grounded with specific parameters, propose behaviors in these regions, and \emph{connector} skills link them by solving boundary value problems. This allows \mosaic{} to discover non-obvious enabling steps, e.g., repositioning an object to allow a feasible grasp. However, every candidate generator and connector is simulated before entering the search graph, whether or not it contributes to the solution. Over 80\% of \mosaic{}'s planning time is spent forward-simulating skills to compute their outcome (Fig.~\ref{fig:mosaic-time}).

To address this computational challenge, we draw on ideas from lazy search \cite{bohlin2000lazyprm, dellin2016lazysp}. Lazy search defers edge evaluation, treating each edge as a hypothesis until it must be resolved. It assumes, however, that the vertices will strictly either be valid or invalid, and that invalidating an edge leaves the rest of the graph unaffected. In \mosaic{}, vertices and edges are both skills, and the outcome of a skill is known only after it is simulated, violating the assumption that lazy search makes. 
In this work, we introduce \lmosaic{}, which extends lazy search to graphs whose vertices, not only edges, are hypotheses. It interleaves two world models: a cheap learned skill-effect model, which predicts the outcome of a skill, and an expensive high-fidelity model, here the physics simulator. \lmosaic{} builds its graph from the cheap model's predictions, queries the expensive model for skills along a promising candidate solution path, and, when a hypothesis is invalidated, repairs the graph locally, keeping the state actually produced and every unaffected hypothesis rather than discarding all of them. Only fully simulated grounded skills ever appear in the returned plan. This crucially preserves physics simulation-based validation, helping alleviate the risk of unexpected contact in safety-critical tasks.

Our contributions are twofold:
\begin{itemize}
    \item We introduce \lmosaic{}, which extends lazy search to skill-centric planning, where deferring a skill also defers the state it produces, so that vertices as well as edges are hypotheses and an invalidated prediction affects a subtree rather than a single edge.
    \item We show how a cheap learned world model can drive this search, deciding which skills are simulated and in what order, while certification guarantees that its accuracy affects only planning efficiency, never the validity of a returned plan with respect to the high-fidelity model.
\end{itemize}

%% file: sections/related.tex
\section{Related Work}
\label{sec:related}

\lmosaic{} draws on three lines of work: skill-based long-horizon planning, lazy search and planning with models of different fidelity, and learned models of action feasibility and effects.

\textbf{Skill-based long-horizon planning.} Task and motion planning (TAMP) \cite{garrett2021integrated} interleaves symbolic search with motion planning, but relies on symbolic abstractions of the preconditions and effects of each skill that are either specified by hand \cite{crosby2016skiros} or learned from data \cite{konidaris2018skills}. Skill-chaining methods rooted in the options framework \cite{sutton1999options, konidaris2009skillchaining} express preconditions and effects as initiation sets and termination conditions in the state space rather than as symbolic predicates, and grow chains incrementally from the start or goal state so that each skill terminates where the next can be initiated. Skills whose initiation sets lie far from the explored frontier are therefore hard to discover. Planning directly with a physics simulator as the transition model dates back to kinodynamic planners such as KPIECE \cite{sucan2009kpiece}, which propagate the system through the simulator and mitigate its cost by steering where the search tree grows. \mosaic{} \cite{mosaic} introduced a different paradigm in which the skills themselves drive the search: generator skills propose configurations and trajectories wherever they are likely to succeed, connector skills join them into a solution, and a physics simulator supplies the effects, so that no symbolic abstractions or hand-specified preconditions are required. Its planning time, however, is dominated by the simulator, since every skill it considers is rolled out before it enters the graph. \lmosaic{} retains this structure and changes when, and how often, the simulator is invoked.

\textbf{Lazy search and models of different fidelity.} When edge evaluation dominates the cost of search, lazy planners defer evaluating an edge until the search has evidence that it is worth the cost. The degree of deferral ranges from evaluating an edge as soon as its successor is expanded \cite{cohen2014lwastar} to evaluating only edges on the current shortest path \cite{bohlin2000lazyprm, dellin2016lazysp}, with intermediate schedules in between \cite{mandalika2019gls} and known conditions under which laziness is beneficial \cite{haghtalab2018lazysp}. 
PACHS \cite{yang2025pachs} defers the expensive rollout of an edge, using a learned critic to decide which edges are worth evaluating. The edge to evaluate next can also be chosen from a learned or inferred prior on edge validity \cite{bhardwaj2019experience, hou2020posterior}. Closest to our setting, \citet{saleem2020selective} plan among movable objects with two models of different fidelity, reserving physics simulation for the actions that require it and using cheap analytic models for the rest.
In all of these, it is assumed that the vertices and edges will only be valid or invalid, so an invalidated element leaves the rest of the graph unaffected. In our setting, deferring a skill also leaves the state it produces unknown, so the contents of vertices are hypothesized as well, and a differing hypothesis affects everything built on it. This calls for a repair mechanism that lazy search does not provide.

\textbf{Learned models of feasibility and effects.} Learned models have been used to replace costly checks, as in the proxy collision checkers of Fastron \cite{das2017fastron}, or, in TAMP, as predictors of the feasibility of a single skill \cite{wells2019feasibility, wang2021learning} or of an abstract plan \cite{yang2023piginet} that let the planner prune candidates before refining them. All of these estimate validity alone, since the state after the action is either known geometrically or not needed. In contact-rich manipulation the effect of a skill on the objects in the scene is precisely the unknown. Learned dynamics models address this, whether planned over directly \cite{chua2018pets, hafner2023dreamerv3} or made object-centric for manipulation \cite{driess2022compositional}, and recent large-scale world models plan in the space of generated video \cite{du2023vlp}. Closest to the model we learn, \citet{liang2022search} search over transitions produced by learned per-skill effect models, and STAP \cite{agia2023stap} chains skills through learned per-skill dynamics models, scoring a sequence by the product of learned skill value functions. In all of these, the output of the learned model is used without verification, as planning transitions or as the training signal for a policy. This can be hazardous in safety-critical tasks such as those in industrial, semi-structured environments, where it is safer to use the most accurate models available (i.e., physics simulation) to alleviate risk of unexpected collisions. \lmosaic{} instead treats the learned model as the cheap one of two world models: its predictions guide a deliberate, non-sequential search over skills and decide which of them are worth a query to the expensive model, the physics simulator, and every trajectory in the returned plan has been validated by the latter.

%% file: sections/background.tex
\section{Algorithmic Background}
\label{sec:background}

\begin{figure}[t]
    \centering
    \includegraphics[width=0.99\linewidth]{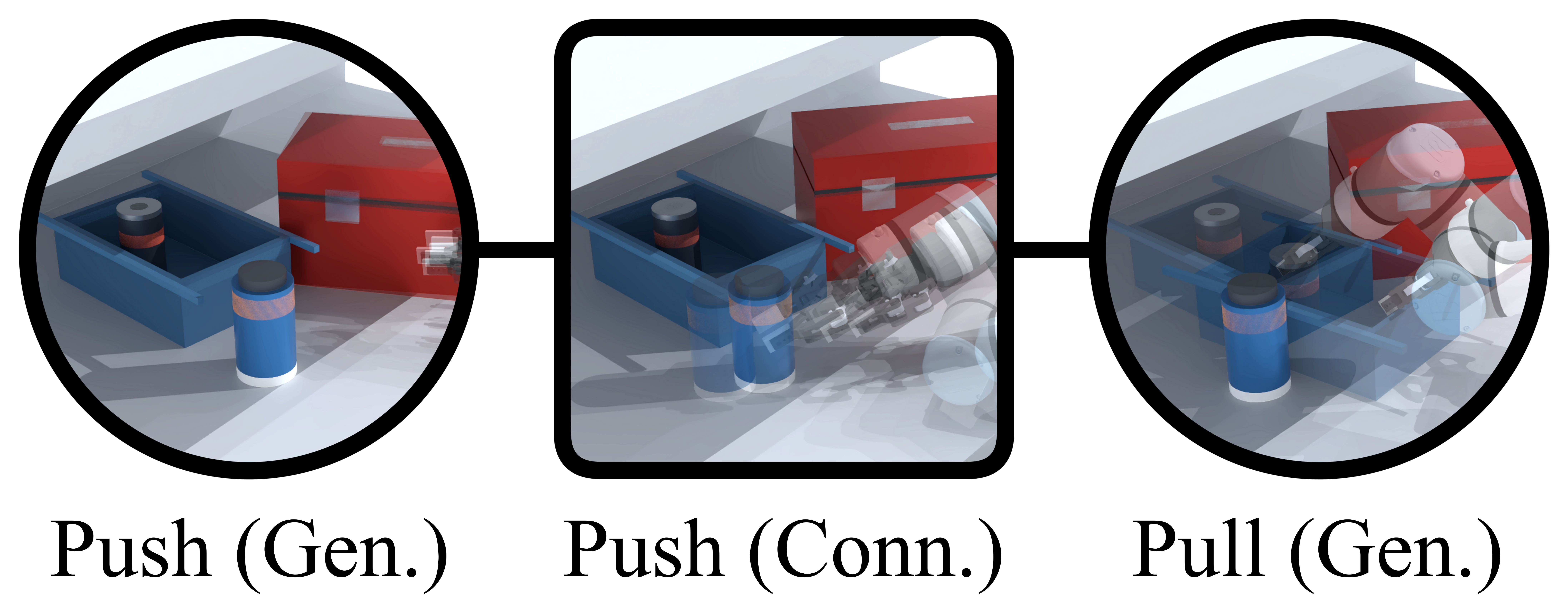}
    \caption{Generators and connectors as in \mosaic{}. Push (Generator) finds a push to move the toolbox. Pull (Generator) finds a trajectory that pulls the bin forward, but starts from a world state where the can is to the left. Push (Connector) bridges the boundary between the two generator skills (the end state of the Push generator and the start state of the Pull generator) by moving the can aside.}
    \label{fig:gen_conn}
\end{figure}

In this section we briefly review the skill-centric planning problem and the components of \mosaic{} \cite{mosaic} on which \lmosaic{} builds. 

Let $\mathcal{X} = \mathcal{Q}_{\mathcal{R}} \times \mathcal{Q}_{\mathcal{O}_1} \times \dots \times \mathcal{Q}_{\mathcal{O}_k}$ denote the world state space, the product of the robot configuration space $\mathcal{Q}_{\mathcal{R}}$ and the configuration spaces $\mathcal{Q}_{\mathcal{O}_i} \subseteq SE(3)$ of the $k$ movable objects. A \emph{binary condition} $\xi: \mathcal{X} \rightarrow \{0,1\}$ indicates whether a state satisfies a property of interest, e.g., whether it is a goal state, and $\Xi$ denotes the set of binary conditions. A \emph{trajectory} is a mapping $\ve{\tau}: [0,1] \rightarrow \mathcal{X}$, and $\Tau$ denotes the space of all trajectories. A \emph{parameterized skill} $\sigma = (\pi_\sigma, \Theta_\sigma)$ consists of a policy $\pi_\sigma$ (e.g., a learned network, an analytic controller, or a motion planner) and a parameter space $\Theta_\sigma$; executing $\pi_\sigma$ with parameters $\theta \in \Theta_\sigma$ produces a trajectory $\ve{\tau} \in \Tau$. \mosaic{} distinguishes two types of skills (Fig.~\ref{fig:gen_conn}). A \emph{generator} $\mathcal{G}_i: \Theta_{\mathcal{G}_i} \rightarrow \Tau$ produces a trajectory without prescribed start or goal conditions, e.g., a grasp of an object in a sampled pose. A \emph{connector} $\mathcal{C}_j: \Xi \times \Xi \times \Theta_{\mathcal{C}_j} \rightarrow \Tau$ produces a trajectory whose start and end states satisfy given boundary conditions. The boundary conditions are most often equalities to given states $x'$ and $x''$, in which case we write $\mathcal{C}_j: \mathcal{X} \times \mathcal{X} \times \Theta_{\mathcal{C}_j} \rightarrow \Tau$, but they may also be the goal condition itself. The \emph{skill library} $\Sigma = \mathcal{G} \cup \mathcal{C}$ is the set of skills available to the planner, where $\mathcal{G} = \{\mathcal{G}_i\}_{i}$ and $\mathcal{C} = \{\mathcal{C}_j\}_{j}$ denote the sets of generators and connectors, respectively.
Given a start state $x_{\text{start}}$, a goal condition $\xi_{\text{goal}} \in \Xi$, and the skill library $\Sigma$, the objective is to find a sequence of skills $\sigma_1, \dots, \sigma_N \in \Sigma$ and their parameters $\theta_i \in \Theta_{\sigma_i}$ whose trajectories $\Pi = \{\ve{\tau}_1, \dots, \ve{\tau}_N\}$ satisfy $\ve{\tau}_1(0) = x_{\text{start}}$, $\ve{\tau}_i(1) = \ve{\tau}_{i+1}(0)$, and $\xi_{\text{goal}}(\ve{\tau}_N(1)) = 1$. That is, a solution is a sequence of parameterized skills that connect continuously and lead from the start state to a state satisfying the goal condition.

\mosaic{} is a non-sequential planner---rather than extending a partial solution forward from the start state, it constructs a directed multigraph, the \emph{mosaic graph}, whose vertices are grounded generator skills, i.e., generators instantiated with specific parameters wherever they are likely to succeed, and whose edges are grounded connector skills joining the end state of one vertex to the start state of another. The start state is represented by a \emph{singleton vertex}, a vertex whose trajectory consists of a single state, so that connectors can attach to it as to any other vertex. An \emph{oracle} orchestrates the construction. At each iteration it selects a skill to invoke, balancing each skill's observed success rate against how often it has been invoked, and samples the skill's parameters. For a connector, it also selects the boundary conditions: either a vertex and its nearest neighbor, favoring vertices already connected to the start or to the goal, or a vertex together with $\xi_{\text{goal}}$, so that the connector attempts to reach a goal state directly from that vertex. The parameterized skill is executed in the physics simulator, and it is added to the graph only if it produces a valid trajectory. Planning terminates once the graph contains a path from $x_{\text{start}}$ to a state satisfying $\xi_{\text{goal}}$. Simulation is essential in this construction--whether a push moves an object as intended, and how it disturbs the objects around it, depends on contact dynamics that cannot be resolved without executing the skill. The graph is therefore valid relative to the simulator by construction, but it grows slowly, since every vertex and every edge requires a computationally expensive rollout in the simulation.
 
Throughout this paper, we refer to any function that characterizes the outcome of executing a parameterized skill from a given state as a \emph{world model}. Given a state $x \in \mathcal{X}$ and a skill $\sigma \in \Sigma$ grounded with parameters $\theta \in \Theta_\sigma$, let an execution outcome consist of its success $o \in \{0,1\}$ and resulting state $x' \in \mathcal{X}$. We denote the expensive high-fidelity model, in our case the physics simulator, by $M_\mathrm{expensive}$ and the cheap learned skill-effect model introduced in Sec.~\ref{sec:world-model} by $M_{\mathrm{cheap}}$. $M_\mathrm{expensive}$ realizes the skill through simulation, producing a trajectory $\tau$ and outcome $o$, with resulting state $x' = \tau(1)$ for a successful execution; $M_\mathrm{cheap}$ instead predicts a success probability $p$, resulting state $\hat{x}$, and uncertainty $u$. Like \mosaic{}, we treat $M_\mathrm{expensive}$ as ground truth: a trajectory is valid if and only if $M_\mathrm{expensive}$ produces it, and validity throughout the paper is defined relative to $M_\mathrm{expensive}$.

%% file: sections/method.tex
\section{\lmosaic{}}
\label{sec:method}

\mosaic{} discovers physically validated skill sequences by rolling out, in a physics simulator, every parameterized skill it considers, whether or not that skill ends up in the final solution. It thus relies on a single, expensive world model for everything it learns about a skill. \lmosaic{} keeps the generators, connectors, and oracle of \mosaic{}, but alleviates its dominant cost---simulating every skill it considers---by introducing a cheap world model that decides which skills are worth a query to the expensive one. Instead of simulating a skill as soon as it is proposed, \lmosaic{} first queries the cheap learned skill-effect model for the outcome of the skill, records the prediction in a separate \emph{optimistic graph}, and queries the simulator selectively, prioritizing the most promising skills. We call this step \emph{certification}: the skill is rolled out in the simulator and, if it succeeds, moves from the optimistic graph to the certified graph. A prediction of the cheap model may be wrong, however: the skill may fail, or it may succeed but produce a different outcome than predicted. Either way, every skill planned downstream of the prediction is affected, so the algorithm must also repair the optimistic graph whenever certification reveals such a mismatch. \lmosaic{} therefore maintains two graphs (Fig.~\ref{fig:optimistic_certified_graphs}), one per world model:
\begin{itemize}
    \item the \textbf{certified graph} $\gcert$, a mosaic graph in the sense of \cite{mosaic}, whose vertices and edges are parameterized skills whose trajectories have been realized in the physics simulator, and
    \item the \textbf{optimistic graph} $\gopt$, whose elements are \emph{proposals} $\rho = (\sigma, \theta, \hat{x}, p, u)$: a skill $\sigma$ with parameters $\theta$, together with the predicted end state $\hat{x}$, predicted success probability $p$, and predictive uncertainty $u$ returned by the cheap model. We call this graph \textit{optimistic} because it provisionally treats predicted skill outcomes as realizable until certification shows otherwise.
\end{itemize}
A generator proposal is a predicted vertex; its source state is either sampled independently, as in \mosaic{}, or taken from the end state of an existing vertex in either graph, certified or predicted. A connector proposal is a predicted edge from the end state of one vertex to the source state of another.\footnote{Strictly, a connector proposal is an edge between two boundary conditions, which may also be the goal condition. As in \mosaic{}, the conditions are most often equalities to states, so in the following we refer to a connector's start and end states.} A path through $\gcert \cup \gopt$ is therefore only a hypothesis about a solution---a proposal may build on the predicted end state of another, so if any prediction along the path is wrong, every proposal that depends on it is affected. \lmosaic{} turns hypotheses into plans by certifying, at each step, a proposal whose predecessors are all certified, so that each skill is simulated from a realized state rather than a predicted one. A path is returned only when all of its elements belong to $\gcert$. In one iteration over the graphs of Fig.~\ref{fig:optimistic_certified_graphs}, the oracle proposes a batch of skills, the cheap model scores them and they enter $\gopt$, a shortest-path query over $\gcert \cup \gopt$ selects a candidate path, and the first uncertified proposal on it is simulated from its realized predecessor. If the outcome matches the prediction, certification continues along the path; otherwise $\rho$ is resolved and only dependent proposals inconsistent with the realized outcome are removed, while the realized state is kept in $\gcert$ for the next iteration to build on. Alg.~\ref{alg:lazy-mosaic} presents the main loop and Alg.~\ref{alg:lazy-mosaic-validate} the certification step; both are detailed below.

\subsection{Skill-Effect World Model}
\label{sec:world-model}

\begin{figure}[t]
    \centering
    \includegraphics[width=0.99\linewidth]{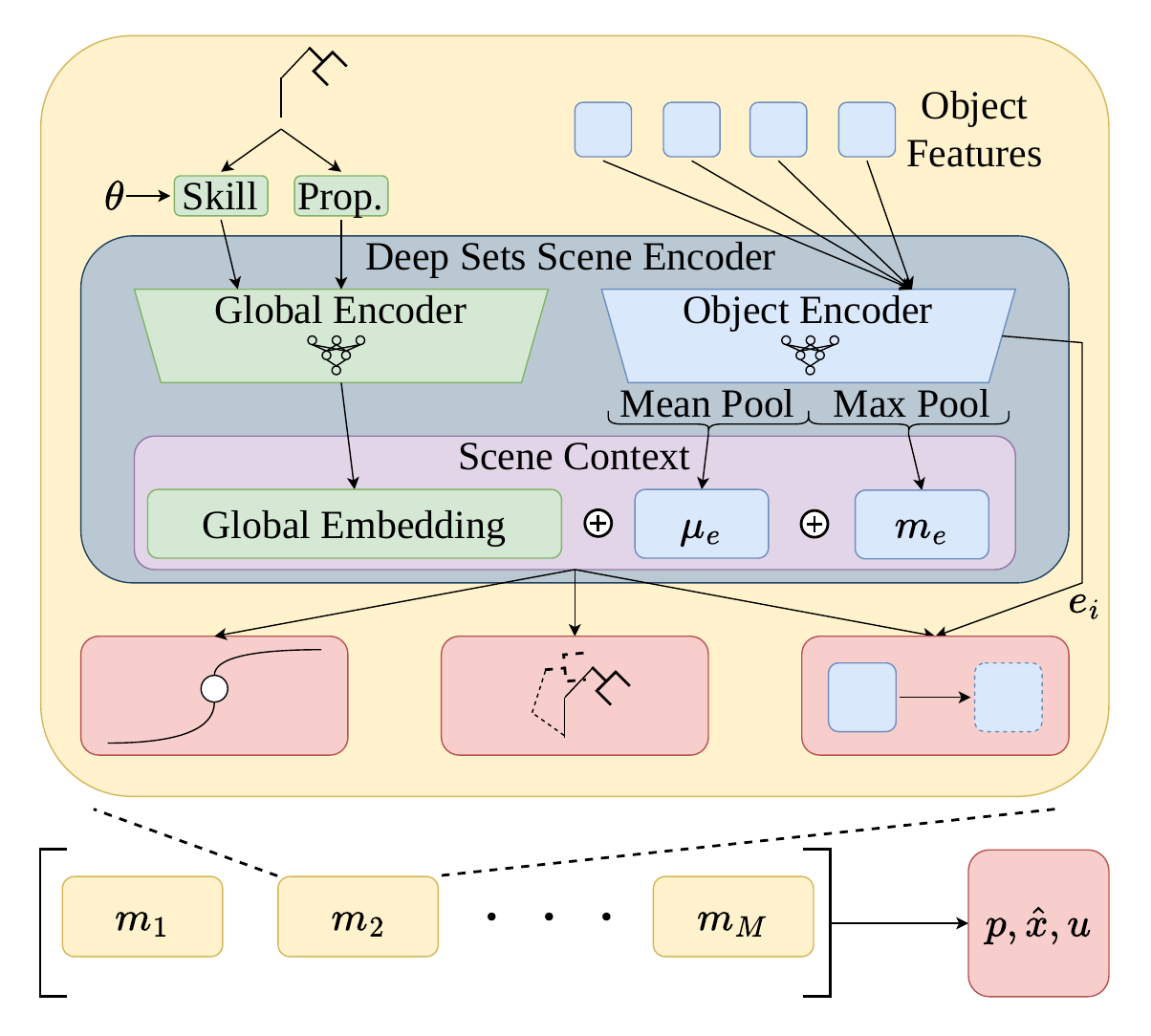}
    \caption{Architecture of the cheap skill-effect world model $M_{\mathrm{cheap}}$. A Deep Sets encoder combines the proposal features, i.e., the skill and the robot's proprioceptive state (Prop.), with per-object features; output heads predict success, change in robot configuration, and object state residuals. An ensemble of $M$ models provides the uncertainty from prediction disagreement.}
    \label{fig:world_model}
\end{figure}

Given a skill $\sigma$ with parameters $\theta$, a source state $x$, and, for connectors, a target state $x'$, the cheap world model $M_{\mathrm{cheap}}$ predicts the outcome of executing the skill from $x$: a probability $p \in [0,1]$ that it produces a valid trajectory achieving its local objective, and a predicted end state $\hat{x} \in \mathcal{X}$. A single model serves the entire skill library and any number of objects (Fig.~\ref{fig:world_model}). Its input has two parts. The \emph{proposal features} describe the queried skill (its identity and parameters $\theta$) and the robot's proprioceptive state (its configuration and end-effector pose in $x$ and, for connectors, its target configuration in $x'$). The \emph{object features} describe each movable object by its pose, geometry, pose relative to the end effector, indicators marking the task target and the manipulated object, and, for connectors, its target pose in $x'$. Following Deep Sets \cite{zaheer2017deepsets}, a shared encoder embeds each object, the object embeddings are pooled, and the result is combined with the embedding of the proposal features into a permutation-invariant scene context. From this context, a success head outputs $p$ and residual heads predict the change in robot configuration and a pose residual for every object. These residuals are relative to a \emph{reference state}, which for a connector is the target state and for a generator is the sampled source state. We train an ensemble of $M$ independently initialized models \cite{lakshminarayanan2017ensembles} and average their outputs. The uncertainty $u$ measures how much the models in the ensemble disagree about where the objects end up. For each object we compute the standard deviation of its predicted position over the models, and $u$ is the largest of these values. Training details are given in Sec.~\ref{sec:experiments}.

\SetKwFunction{SampleParameters}{SampleParameters}
\SetKwFunction{Certify}{Certify}
\SetKw{KwAnd}{and}
\SetKw{KwOr}{or}
\SetKw{KwWith}{with}

\begin{algorithm}[t]
    \scriptsize
    \caption{\lmosaic{}}
    \label{alg:lazy-mosaic}
    \KwIn{Start state $x_{\mathrm{start}} \in \mathcal{X}$ \newline
          Goal condition $\xi_{\mathrm{goal}}: \mathcal{X} \rightarrow \{0,1\}$ \newline
          Skill library $\Sigma$ and oracle $O$ \newline
          Cheap world model $M_{\mathrm{cheap}}$\newline
          Expensive world model $M_{\mathrm{expensive}}$ (physics simulator) \newline
          Batch size $K$, cost weights $w_1, w_2, w_3$, stall threshold $N_{\mathrm{stall}}$, random-certification probability $p_{\mathrm{rand}}>0$}
    \KwOut{Physics-certified path $\Pi$, or failure}
    \vspace{4pt}
    \tcp{\color{orange} \scriptsize $\gcert$ contains certified skills only, \newline
    $\gopt$ predictions only}
    $\gcert \gets \{x_{\mathrm{start}}\},\; \gopt \gets \emptyset$\;
    \While{planning budget remains}{
        \If{$\gcert$ contains a path $\Pi$ from $x_{\mathrm{start}}$ to a state satisfying $\xi_{\mathrm{goal}}$\label{alg1:certpath}}{
            \KwRet $\Pi$\label{alg1:return}\;
        }
        \lIf{$\gopt$ has an eligible proposal \KwAnd
      $\mathrm{rand}() < p_{\mathrm{rand}}$}{
    $\Certify(\gopt.\mathrm{randomEligible}(), \gcert, \gopt, M_{\mathrm{expensive}})$; \Continue
}
        \eIf{$\gcert$ has not grown for $N_{\mathrm{stall}}$ iterations \KwAnd $\gopt$ has an eligible proposal\label{alg1:stall}}{
            \tcp{\color{orange} \scriptsize Stall rule. Returns the lowest-cost proposal among those whose predecessors are all certified}
            $\rho \gets \gopt.\mathrm{topEligible}()$\;
            $\Certify(\rho, \gcert, \gopt, M_{\mathrm{expensive}})$\;
            continue \label{alg1:stallend}\;
        }{
            \tcp{\color{orange} \scriptsize Optimistic expansion}
            $\sigma \gets O.\ChooseSkill(\Sigma, \gcert \cup \gopt)$\label{alg1:choose}\;
            $\theta_1, \dots, \theta_K \gets \SampleParameters(\sigma, \gcert \cup \gopt, K)$\label{alg1:sample}\;
            $\{(\hat{x}_i, p_i, u_i)\}_{i=1}^{K} \gets M_{\mathrm{cheap}}(\sigma, \{\theta_i\}_{i=1}^{K})$ \label{alg1:predict}\tcp*{\color{orange} \scriptsize Batch}
            insert proposals $\rho_i = (\sigma, \theta_i, \hat{x}_i, p_i, u_i)$ into $\gopt$ with costs $c^{\mathrm{opt}}_i$ from Eq.~\eqref{eq:opt-cost}\label{alg1:insert}\;
            $\Pi \gets \ShortestPath(x_{\mathrm{start}}, \xi_{\mathrm{goal}}, \gcert \cup \gopt)$\label{alg1:optpath}\;
        }
        \lIf{$\Pi = \emptyset$}{\Continue}
        \tcp{\color{orange} \scriptsize Certify $\Pi$ one proposal at a time; consistent is false once a realized outcome differs from its prediction}
        $\mathrm{consistent} \gets \mathrm{true}$\label{alg1:intact}\;
        \While{$\mathrm{consistent}$ \KwAnd $\Pi$ has an uncertified proposal\label{alg1:certloop}}{
            $\rho \gets$ first uncertified proposal on $\Pi$\label{alg1:first}\;
            $\mathrm{consistent} \gets \Certify(\rho, \gcert, \gopt, M_{\mathrm{expensive}})$ \label{alg1:validate}\tcp*{\color{orange} \scriptsize Alg.~\ref{alg:lazy-mosaic-validate}}
        }
    }
    \KwRet failure\;
\end{algorithm}

\begin{algorithm}[t]
    \scriptsize
    \caption{Certify}
    \label{alg:lazy-mosaic-validate}
    \KwIn{Proposal $\rho = (\sigma, \theta, \hat{x}, p, u)$ \newline
          Graphs $\gcert$ and $\gopt$ \newline
          Expensive world model $M_{\mathrm{expensive}}$ \newline
          Tolerances $\epsilon_p, \epsilon_R, \epsilon_q$}
    \KwOut{$\mathrm{true}$ iff the realized outcome is consistent with the prediction}
    \vspace{4pt}
    \tcp{\color{orange} \scriptsize $\Match(x, x')$: per-object position error $\le \epsilon_p$, orientation error $\le \epsilon_R$ up to object symmetries, arm-joint error $\le \epsilon_q$}
    $x \gets$ realized end state of the predecessor of $\rho$\;
    $(\ve{\tau}, o) \gets M_{\mathrm{expensive}}(\sigma, \theta, x)$ \label{alg2:rollout}\tcp*{\color{orange} \scriptsize expensive query}
    \If{$o$ is a failure}{
        \If{$\sigma \in \mathcal{G}$}{
            \tcp{\color{orange} \scriptsize proposals built on $\rho$, including uncertified connectors into it}
            $\gopt.\mathrm{invalidateDependents}(\rho)$\label{alg2:invdep}\;
            \tcp{\color{orange} \scriptsize certified connectors into $\rho$ keep their trajectory; each now ends at a singleton vertex}
            $\gcert.\mathrm{toSingleton}(\text{connectors into } \rho)$\label{alg2:incoming}\;
        }
        $\gopt.\mathrm{remove}(\rho)$\label{alg2:invc}\;
        \KwRet $\mathrm{false}$\label{alg2:failret}\;
    }
    $\gcert.\mathrm{insert}(\sigma, \ve{\tau}, c^{\mathrm{cert}} = w_1 a(\sigma, \theta, \ve{\tau}(1)))$ \label{alg2:insert}\tcp*{\color{orange} \scriptsize Eq.~\eqref{eq:opt-cost}, $p{=}1$, $u{=}0$}
    $m \gets \Match(\ve{\tau}(1), \hat{x})$\label{alg2:match}\;
    \uIf{$\sigma \in \mathcal{C}$\label{alg2:conn}}{
        $g \gets$ vertex that $\rho$ was intended to reach\;
        \If{$\neg m \;\vee\; \neg\Match(\ve{\tau}(1), g.\mathrm{start})$}{
            $\gcert.\mathrm{toSingleton}((\sigma, \ve{\tau}))$\; $\gopt.\mathrm{removeLink}(\rho, g)$\;
            $\gopt.\mathrm{remove}(\rho)$\label{alg2:invc}\;
            \KwRet $\mathrm{false}$\label{alg2:connret}\;
        }
    }
    \ElseIf{$\neg m$\label{alg2:gen}}{
        \lFor{$d \in \gopt.\mathrm{dependents}(\rho)$ \KwWith $\neg\Match(\ve{\tau}(1), d.\mathrm{start})$}{$\gopt.\mathrm{invalidate}(d)$}
        $\gopt.\mathrm{remove}(\rho)$\label{alg2:invc}\;
        \KwRet $\mathrm{false}$\label{alg2:genret}\;
    }
    $\gopt.\mathrm{remove}(\rho)$\label{alg2:invc}\;
    \KwRet $\mathrm{true}$\label{alg2:true}\;
\end{algorithm}

\subsection{Optimistic Expansion}
\label{sec:expansion}

Each iteration of Alg.~\ref{alg:lazy-mosaic} first checks whether $\gcert$ already contains a start-to-goal path and returns it if so (Lines \ref{alg1:certpath}--\ref{alg1:return}). Otherwise the oracle selects a skill as in \mosaic{}, using per-skill success statistics that are updated from certified outcomes only (Line \ref{alg1:choose}), and samples a batch of $K$ parameter sets for it (Line \ref{alg1:sample}). For generators, if a source state is required, it is either sampled at random, or taken from the end state of a vertex in $\gcert$ or $\gopt$, so that a generator can also be seeded from a predicted intermediate state, e.g., the shelf after the toolbox has been pushed aside. For connectors, we select either a vertex and its nearest neighbor, favoring vertices connected to the start or goal, or a vertex together with $\xi_{\text{goal}}$ (Sec.~\ref{sec:background}). The cheap world model scores the batch (Line \ref{alg1:predict}), and each proposal is assigned the optimistic cost
\begin{equation}
    c^{\mathrm{opt}} = w_1\, a(\sigma, \theta, \hat{x}) + w_2\, s + w_3\, u,
    \label{eq:opt-cost}
\end{equation}
where $a$ is the cost of executing the skill (e.g., a unit cost per skill, or the trajectory length), $s = -\log(\tilde{p})$ with $\tilde{p}=\max{(p,10^{-6})}$ for numerical stability, $w_2 > 0$ penalizes proposals the model expects to fail, and $w_3 > 0$ penalizes proposals whose effect the model is uncertain about. The proposals are inserted into $\gopt$ with their costs (Line \ref{alg1:insert}), and a shortest-path query over $\gcert \cup \gopt$ returns the current best candidate solution if one exists (Line \ref{alg1:optpath}). We say that a proposal is \emph{eligible} if all of its predecessors have been certified, i.e., its source state has been realized in the simulator. Only eligible proposals are certified, so a proposal deep in a predicted chain waits until the chain below it has been realized. If $\gcert$ has not grown for $N_{\mathrm{stall}}$ iterations, which indicates that the optimistic search repeatedly selects paths whose first proposals fail, \lmosaic{} certifies the lowest-cost eligible proposal directly, whether or not it lies on a complete optimistic path (Lines \ref{alg1:stall}--\ref{alg1:stallend}). This rule keeps the certified graph growing when optimistic search stalls. At each iteration, with probability $p_\mathrm{rand}>0$, \lmosaic{} instead certifies a uniformly random eligible proposal. This preserves the probabilistic completeness of \mosaic{}, that only requires that every skill be sampled and simulated with nonzero probability.

\subsection{Certification and Repair}
\label{sec:certification}

Given a candidate path $\Pi$, \lmosaic{} selects its first uncertified proposal $\rho$ (Alg.~\ref{alg:lazy-mosaic}, Lines \ref{alg1:intact}--\ref{alg1:validate}) and rolls out its skill in the simulator from the \emph{realized} predecessor state (Alg.~\ref{alg:lazy-mosaic-validate}, Line \ref{alg2:rollout}). If the rollout fails, no certified counterpart is added to $\gcert$. For a failed connector, only that proposal is lost; for a failed generator, every proposal that depended on its predicted state is invalidated (Line \ref{alg2:invdep}), and connectors already certified \emph{into} its start state are re-attached to singleton vertices at their realized end states (Line \ref{alg2:incoming}), so that the physically realized states they reach remain available for later connections.

If the rollout succeeds, the realized trajectory $\ve{\tau}$ is inserted into $\gcert$ at its certified cost, namely Eq.~\eqref{eq:opt-cost} with $p = 1$ and $u = 0$ (Line \ref{alg2:insert}). If the realized end state $\ve{\tau}(1)$ matches the prediction $\hat{x}$ within per-object position and orientation tolerances $(\epsilon_p, \epsilon_R)$ and a joint-space tolerance $\epsilon_q$ (Line \ref{alg2:match}), the dependents of $\rho$ remain in $\gopt$ and certification proceeds to the next proposal along $\Pi$ (Line \ref{alg2:true}, and Alg.~\ref{alg:lazy-mosaic}, Lines \ref{alg1:certloop}--\ref{alg1:validate}). 
Otherwise $\gopt$ is repaired locally around the mismatch. A connector that was intended to arrive at a generator vertex $g$ but whose realized end state fails to match either its own prediction or the start state of $g$ is re-attached to a singleton vertex at its realized end state and its link into $g$ is removed (Alg.~\ref{alg:lazy-mosaic-validate}, Lines \ref{alg2:conn}--\ref{alg2:connret}); for a generator, each direct dependent whose start state no longer matches the realized state is invalidated, while consistent dependents are retained (Lines \ref{alg2:gen}--\ref{alg2:genret}). In both cases certification along $\Pi$ stops and control returns to the search, which now observes the realized state in $\gcert$ and can propose new skills from it, while predictions elsewhere in $\gopt$ remain usable. Certification always resolves the selected proposal $\rho$, which is removed from $\gopt$ after evaluation, and, on success, replaced by its realized trajectory in $\gcert$. Other optimistic proposals are retained unless the certified outcome invalidates a state on which they depend.

\begin{observation}[Soundness relative to $M_\mathrm{expensive}$]
\label{prop:soundness}
Let $\Pi = (\ve{\tau}_1,\ldots,\ve{\tau}_N)$ be a plan returned
by \lmosaic{}. Every $\ve{\tau}_i$ is produced by the expensive
world model $M_\mathrm{expensive}$, with
$\ve{\tau}_1(0)=x_{\mathrm{start}}$ and
$\ve{\tau}_i(0)=\ve{\tau}_{i-1}(1)$ for $i>1$.
Moreover, $\xi_{\mathrm{goal}}(\ve{\tau}_N(1))=1$.
Thus, no predicted state from $\gopt$ appears in a returned plan.
\end{observation}

\begin{proof}
A path is returned only when it lies entirely in $\gcert$
(Alg.~\ref{alg:lazy-mosaic}, Lines
\ref{alg1:certpath}--\ref{alg1:return}), and an element enters
$\gcert$ only after $M_\mathrm{expensive}$ successfully rolls out its skill
from the realized end state of its certified predecessor, or from
$x_{\mathrm{start}}$ for the first element
(Alg.~\ref{alg:lazy-mosaic-validate}, Line~\ref{alg2:rollout}).
Induction along the returned path therefore gives
$\ve{\tau}_i(0)=\ve{\tau}_{i-1}(1)$ with every $\ve{\tau}_i$ produced
by $M_\mathrm{expensive}$, and the path is returned only once its certified
terminal state satisfies $\xi_{\mathrm{goal}}$. Since validity is
defined relative to $M_\mathrm{expensive}$ (Sec.~\ref{sec:background}), $\Pi$
is a valid plan; predictions and the \textsc{Match} test determine
which proposals are retained, not the states from which certified
trajectories are rolled out, so $M_\mathrm{cheap}$ affects only which
proposals are certified and in which order.
\end{proof}

%% file: sections/experiments.tex
\section{Experimental Analysis}
\label{sec:experiments}

\begin{figure*}[t]
    \centering
    
    \begin{subfigure}[b]{0.32\linewidth}
        \centering
        \includegraphics[width=\linewidth]{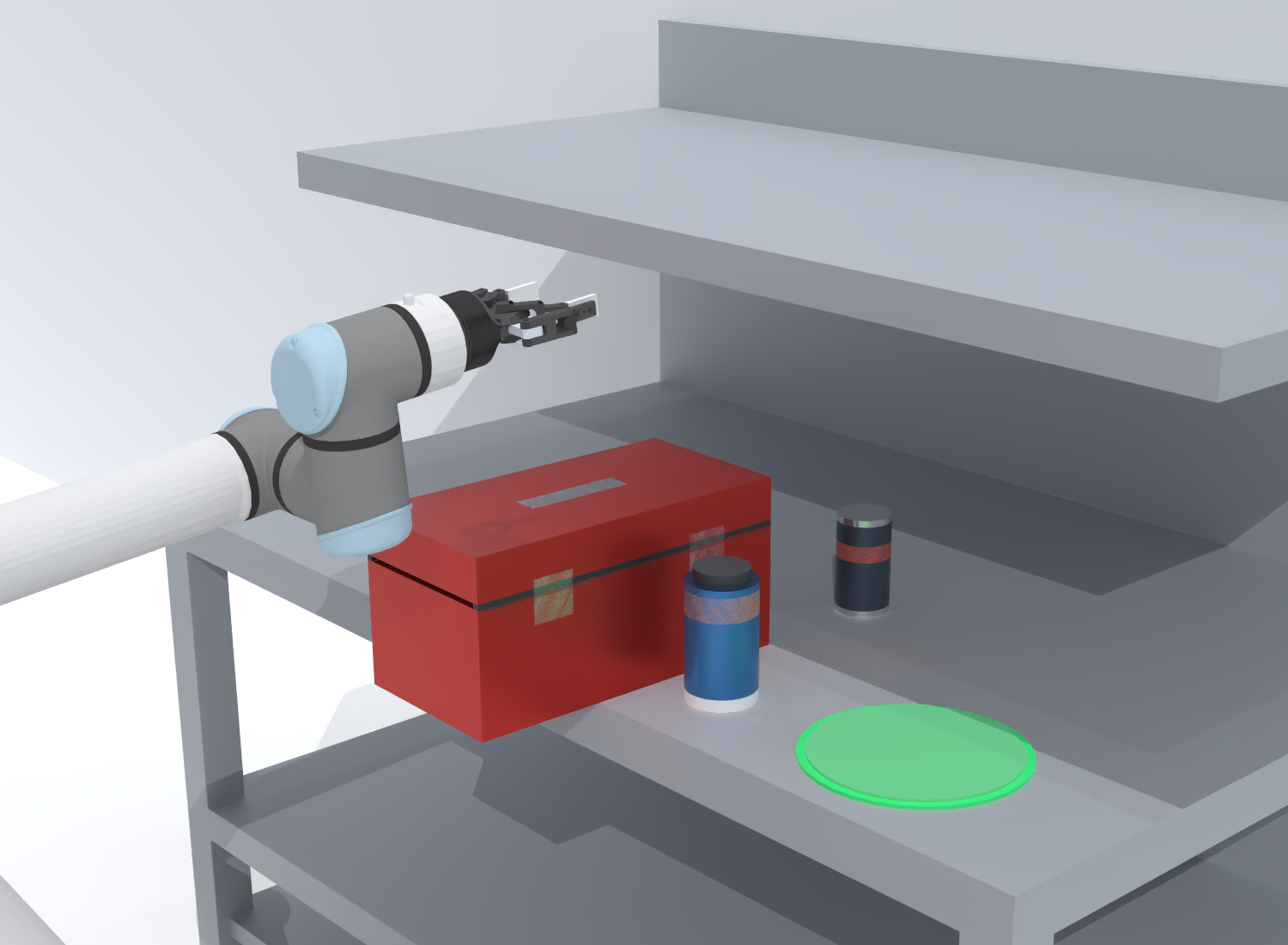}
        \caption{Task 1: Cluttered retrieval.}
        \label{fig:task1}
    \end{subfigure}
    \hfill 
    \begin{subfigure}[b]{0.32\linewidth}
        \centering
        \includegraphics[width=\linewidth]{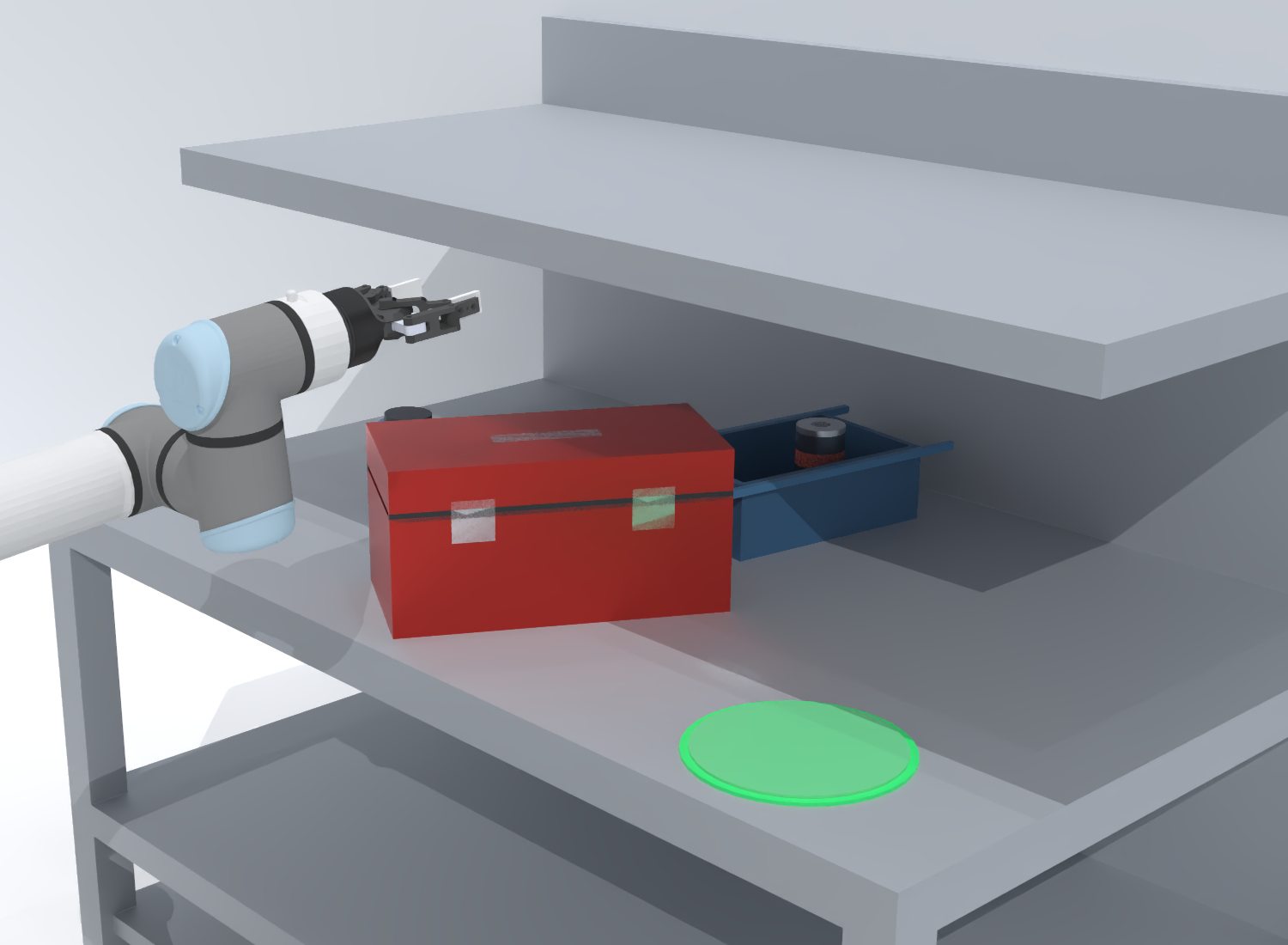}
        \caption{Task 2: Bin manipulation.}
        \label{fig:task2}
    \end{subfigure}
    \hfill 
    \begin{subfigure}[b]{0.32\linewidth}
        \centering
        \includegraphics[width=\linewidth]{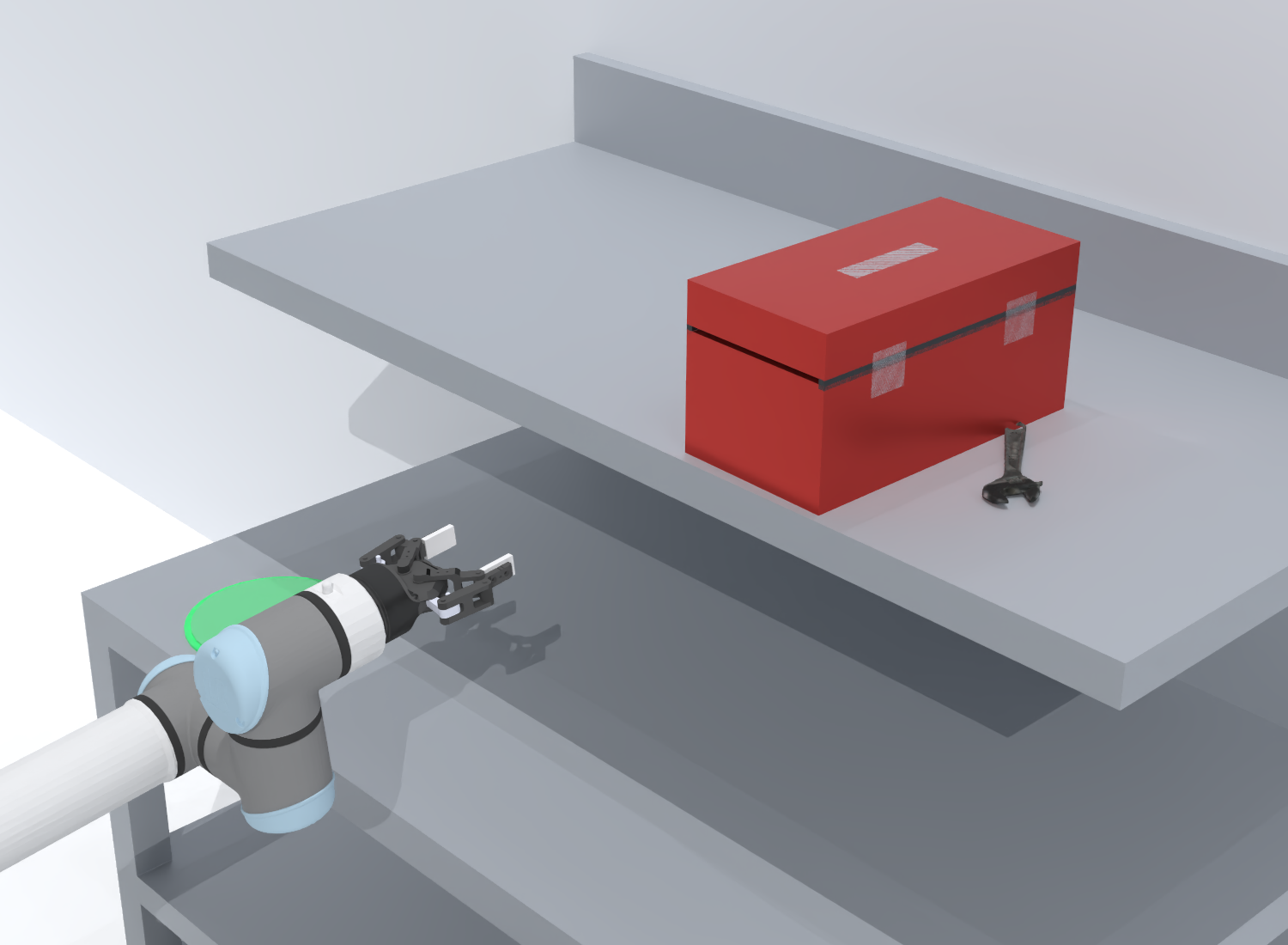}
        \caption{Task 3: Wrench repositioning.}
        \label{fig:task3}
    \end{subfigure}

    \caption{
    Simulation tasks, shown in their hard configurations.
    From left to right: Task~1, cluttered object retrieval; Task~2,
    retrieval requiring manipulation of the containing bin; and Task~3,
    wrench retrieval requiring the wrench to first be repositioned.
    Easy and medium variants progressively remove clutter while preserving
    the task objective. The green region denotes the goal region.
    }
    \label{fig:tasks}
\end{figure*}

We evaluate \lmosaic{} on long-horizon manipulation tasks with generic, imperfect skills to assess its planning efficiency and success rate, the accuracy of the learned model, and the role of physics-based certification. Please refer to the accompanying video for a demonstration of \lmosaic{} on a real robot and in simulation.

\begin{table*}[!t]
    \centering
    \caption{Planning performance over 20 trials per setting. \emph{Time} is the mean planning time in seconds over successful runs only. SR is the success rate and \emph{Len} the mean solution length in skills over the successful runs. Boldface marks the lowest time among methods with nonzero success and the highest nonzero success rate in each setting; ties are all bolded.}
    \label{tab:planning_without_timeouts}
    \scriptsize
    \setlength{\tabcolsep}{2.0pt}
    \renewcommand{\arraystretch}{1.08}
    \resizebox{\textwidth}{!}{%
    \begin{tabular}{@{}l*{8}{ccc}@{}}
        \toprule
        & \multicolumn{9}{c}{\textbf{Task 1}} & \multicolumn{9}{c}{\textbf{Task 2}} & \multicolumn{6}{c}{\textbf{Task 3}} \\
        \cmidrule(lr){2-10}\cmidrule(lr){11-19}\cmidrule(lr){20-25}
        & \multicolumn{3}{c}{Easy} & \multicolumn{3}{c}{Medium} & \multicolumn{3}{c}{Hard}
        & \multicolumn{3}{c}{Easy} & \multicolumn{3}{c}{Medium} & \multicolumn{3}{c}{Hard}
        & \multicolumn{3}{c}{Medium} & \multicolumn{3}{c}{Hard} \\
        \cmidrule(lr){2-4}\cmidrule(lr){5-7}\cmidrule(lr){8-10}\cmidrule(lr){11-13}\cmidrule(lr){14-16}\cmidrule(lr){17-19}\cmidrule(lr){20-22}\cmidrule(lr){23-25}
        Method
        & Time & SR & Len & Time & SR & Len & Time & SR & Len
        & Time & SR & Len & Time & SR & Len & Time & SR & Len
        & Time & SR & Len & Time & SR & Len \\
        \midrule
        \mosaic{}
        & 104.4 & \textbf{20/20} & 2.0 & 273.2 & 16/20 & 3.69 & 418.0 & 7/20 & 5.86 & 144.2 & \textbf{20/20} & 3.75 & 323.6 & 8/20 & 7.63 & 540.5 & 4/20 & 10.25 & 397.2 & 6/20 & 5.67 & N/A & 0/20 & -- \\
        \textbf{\lmosaic{}}
        & 44.1 & \textbf{20/20} & 2.0 & 67.3 & \textbf{19/20} & 3.58 & 175.1 & \textbf{13/20} & 6.23 & 41.0 & \textbf{20/20} & 3.45 & \textbf{121.7} & \textbf{10/20} & 7.8 & \textbf{256.8} & \textbf{7/20} & 10.57 & \textbf{60.3} & \textbf{11/20} & 6.18 & \textbf{584.8} & \textbf{4/20} & 10.5 \\
        Ranked-\mosaic{}
        & 91.6 & \textbf{20/20} & 2.0 & 238.9 & 17/20 & 3.71 & 359.4 & 9/20 & 6.11 & 126.5 & \textbf{20/20} & 3.95 & 294.8 & 8/20 & 7.38 & 501.0 & 5/20 & 11.0 & 326.7 & 7/20 & 5.86 & N/A & 0/20 & -- \\
        \lmosaic{} w/o uncertainty
        & 59.0 & \textbf{20/20} & 2.0 & 97.2 & 18/20 & 3.83 & 232.8 & 11/20 & 6.45 & 53.0 & \textbf{20/20} & 4.0 & 178.9 & 9/20 & 8.11 & 287.8 & 4/20 & 10.75 & 136.5 & 7/20 & 6.29 & 597.1 & 1/20 & 10.0 \\
        WM-\mosaic{}
        & \textbf{1.7} & 18/20 & 2.0 & \textbf{2.4} & 7/20 & 3.86 & \textbf{8.1} & 2/20 & 6.5 & \textbf{3.69} & 6/20 & 4.33 & N/A & 0/20 & -- & N/A & 0/20 & -- & N/A & 0/20 & -- & N/A & 0/20 & -- \\
        \midrule
        Options
        & 114.6 & 17/20 & 2.0 & 472.6 & 5/20 & 3.8 & N/A & 0/20 & -- & 299.4 & 15/20 & 4.53 & 523.3 & 3/20 & 8.67 & N/A & 0/20 & -- & N/A & 0/20 & -- & N/A & 0/20 & -- \\
        Lazy Options
        & 66.9 & 13/20 & 2.0 & 156.7 & 8/20 & 3.88 & 533.3 & 1/20 & 7.0 & 49.1 & 11/20 & 4.82 & N/A & 0/20 & -- & N/A & 0/20 & -- & N/A & 0/20 & -- & N/A & 0/20 & -- \\
        \midrule
        CEM
        & 440.5 & 11/20 & 2.0 & 577.2 & 1/20 & 5.0 & N/A & 0/20 & -- & 559.2 & 4/20 & 5.75 & N/A & 0/20 & -- & N/A & 0/20 & -- & N/A & 0/20 & -- & N/A & 0/20 & -- \\
        WM-CEM
        & 101.2 & 11/20 & 2.0 & 262.8 & 4/20 & 4.75 & N/A & 0/20 & -- & 57.8 & 4/20 & 6.0 & N/A & 0/20 & -- & N/A & 0/20 & -- & N/A & 0/20 & -- & N/A & 0/20 & -- \\
        \midrule
        Roadmap
        & 233.6 & \textbf{20/20} & 2.0 & 572.5 & 6/20 & 4.0 & 593.0 & 1/20 & 7.0 & 319.5 & 16/20 & 4.63 & 567.1 & 5/20 & 8.2 & N/A & 0/20 & -- & N/A & 0/20 & -- & N/A & 0/20 & -- \\
        Lazy Roadmap
        & 83.4 & 19/20 & 2.0 & 271.0 & 9/20 & 4.11 & 471.3 & 3/20 & 6.33 & 173.4 & 12/20 & 4.92 & 324.4 & 1/20 & 9.0 & N/A & 0/20 & -- & N/A & 0/20 & -- & N/A & 0/20 & -- \\
        \bottomrule
    \end{tabular}%
    }
\end{table*}

\begin{table}[!t]
    \centering
    \captionsetup{
        font=scriptsize,
        justification=raggedright,
        singlelinecheck=false
    }
    \setlength{\tabcolsep}{1.1pt}

    \begin{minipage}[t]{0.49\columnwidth}
        \vspace{0pt}
        \centering

        \caption{Mean number of simulator queries per trial, by
        method and difficulty. WM-\mosaic{} makes none.}
        \label{tab:sim-skills}

        \scriptsize
        \resizebox{\linewidth}{!}{%
        \begin{tabular}{@{}lccc@{}}
            \toprule
            Method & Easy & Medium & Hard \\
            \midrule
            \mosaic{} & 46.4 & 67.9 & 134.5 \\
            \lmosaic{} & \textbf{16.9} & \textbf{21.7} & \textbf{41.3} \\
            Ranked-\mosaic{} & 44.3 & 62.9 & 107.9 \\
            \lmosaic{} w/o uncertainty & 19.0 & 25.9 & 50.7 \\
            WM-\mosaic{} & 0 & 0 & 0 \\
            \midrule
            Options & 54.9 & 74.8 & 251.4 \\
            Lazy Options & 23.7 & 32.0 & 99.6 \\
            \midrule
            CEM & 120.2 & 214.2 & 369.8 \\
            WM-CEM & 36.2 & 68.0 & 219.1 \\
            \midrule
            Roadmap & 67.2 & 99.3 & 200.2 \\
            Lazy Roadmap & 35.8 & 45.8 & 84.1 \\
            \bottomrule
        \end{tabular}%
        }

    \end{minipage}
    \hfill
    \begin{minipage}[t]{0.48\columnwidth}
        \vspace{0pt}
        \centering


        \begin{tikzpicture}
        \begin{axis}[
            width=\linewidth,
            height=2.15cm,
            ybar,
            bar width=5.2pt,
            ymin=0,
            ymax=2100,
            symbolic x coords={Easy,Medium,Hard},
            xtick=data,
            xticklabel style={font=\scriptsize,yshift=1pt},
            ytick={0,1000,2000},
            yticklabel style={font=\tiny},
            ylabel={Vertices},
            ylabel style={font=\scriptsize,yshift=-3pt},
            nodes near coords,
            point meta=y,
            every node near coord/.append style={
                font=\tiny,
                inner sep=0.5pt,
                /pgf/number format/fixed,
                /pgf/number format/precision=1
            },
            tick align=outside,
            axis x line*=bottom,
            axis y line*=left,
            axis line style={black!65},
            ymajorgrids,
            grid style={black!12},
            enlarge x limits=0.20,
            scaled y ticks=false,
            clip=false,
        ]
            \addplot[
                draw=cyan!60!black,
                fill=cyan!35
            ] coordinates {
                (Easy,308)
                (Medium,875)
                (Hard,1736)
            };

            \addplot[
                draw=black,
                fill=black!55
            ] coordinates {
                (Easy,8)
                (Medium,13)
                (Hard,17)
            };
        \end{axis}
        \end{tikzpicture}

        \vspace{-1.3ex}

        \begin{tikzpicture}
        \begin{axis}[
            width=\linewidth,
            height=1.95cm,
            ybar,
            bar width=5.2pt,
            ymin=0,
            ymax=720,
            symbolic x coords={Easy,Medium,Hard},
            xtick=data,
            xticklabel style={font=\scriptsize,yshift=1pt},
            ytick={0,300,600},
            yticklabel style={font=\tiny},
            ylabel={Edges},
            ylabel style={font=\scriptsize,yshift=-3pt},
            nodes near coords,
            point meta=y,
            every node near coord/.append style={
                font=\tiny,
                inner sep=0.5pt,
                /pgf/number format/fixed,
                /pgf/number format/precision=1
            },
            tick align=outside,
            axis x line*=bottom,
            axis y line*=left,
            axis line style={black!65},
            ymajorgrids,
            grid style={black!12},
            enlarge x limits=0.20,
            scaled y ticks=false,
            clip=false,
        ]
            \addplot[
                draw=cyan!60!black,
                fill=cyan!35
            ] coordinates {
                (Easy,64)
                (Medium,317)
                (Hard,587)
            };

            \addplot[
                draw=black,
                fill=black!55
            ] coordinates {
                (Easy,4)
                (Medium,5)
                (Hard,8)
            };
        \end{axis}
        \end{tikzpicture}

        \vspace{-0.8ex}

        {\tiny
        \tikz[baseline=-0.5ex]
            \draw[draw=cyan!60!black,fill=cyan!35]
            (0,0) rectangle (0.14,0.09);
        \hspace{0.2em}Optimistic
        \qquad
        \tikz[baseline=-0.5ex]
            \draw[draw=black,fill=black!55]
            (0,0) rectangle (0.14,0.09);
        \hspace{0.2em}Certified
        }

        \vspace{-0.6ex}

        \captionof{figure}{
            Optimistic versus certified graph size for \lmosaic{}.
        }
        \label{fig:opt-cert}


        \vspace{1.2ex}

        \begin{tikzpicture}
            \def\barwidth{4.25}
            \def\barheight{0.32}

            \fill[
                cyan!40,
                draw=cyan!60!black
            ]
            (0,0)
            rectangle
            ({0.834*\barwidth},\barheight);

            \fill[
                black!20,
                draw=black!45
            ]
            ({0.834*\barwidth},0)
            rectangle
            ({0.915*\barwidth},\barheight);

            \fill[
                black!55,
                draw=black!70
            ]
            ({0.915*\barwidth},0)
            rectangle
            (\barwidth,\barheight);

            \draw[black!65]
            (0,0) rectangle (\barwidth,\barheight);

            \node[
                font=\tiny\bfseries
            ]
            at ({0.417*\barwidth},{0.5*\barheight})
            {83.4\%};

        \end{tikzpicture}

        \vspace{0.4ex}

        {\tiny
        \tikz[baseline=-0.5ex]
            \draw[draw=cyan!60!black,fill=cyan!40]
            (0,0) rectangle (0.14,0.09);
        \hspace{0.15em}Physics
        \qquad
        \tikz[baseline=-0.5ex]
            \draw[draw=black!45,fill=black!20]
            (0,0) rectangle (0.14,0.09);
        \hspace{0.15em}IK/motion planning
        \qquad
        \tikz[baseline=-0.5ex]
            \draw[draw=black!70,fill=black!55]
            (0,0) rectangle (0.14,0.09);
        \hspace{0.15em}Other
        }

        \vspace{-0.4ex}

        \captionof{figure}{
            \mosaic{} planning-time breakdown:
            physics simulation accounts for 83.4\%,
            IK/motion planning for 8.1\%, and other computation for 8.5\%.
        }
        \label{fig:mosaic-time}

    \end{minipage}

\end{table}

\subsection{Experimental Setup}

We use a UR10e with a parallel-jaw gripper in the three shelf-manipulation tasks shown in Fig.~\ref{fig:tasks}, simulated in ManiSkill3 \cite{gu2023maniskill2} with SAPIEN \cite{Xiang_2020_SAPIEN}. We target a semi-structured manipulation environment where the workspace geometry and object types are known across tasks, but object presence and poses vary between instances. Tasks 1 and 2 each have easy, medium, and hard variants, Task 3 has no easy variant: even without clutter, retrieving the wrench requires first repositioning it to the shelf edge before it can be grasped. All planners are evaluated on the same 20 randomized configurations per setting, with a 600 s planning timeout. No task-specific knowledge is given beyond the goal condition. We report success rate, planning time excluding execution, solution length in skills, and the number of simulator queries.

\subsubsection{Skill Library} All planners are given access to the same library of five learned and analytical object-centric skills. Every collision-free arm motion within these skills is generated with SRMP \cite{mishani2025srmp}. \emph{Pick} (generator) computes grasps from object geometry, plans a motion to a pre-grasp pose, and executes a screw-based approach, grasp, and retract. \emph{Place} (connector) similarly plans a motion to a pre-place pose and releases a held object in a specified region. \emph{Move} (connector) plans motions between robot configurations, including while transporting a held object. Note there is no single pick-and-place skill; the planner must discover that Pick and Place have to be composed. \emph{Push} (generator and connector) and \emph{Pull} (generator) use flow-matching policies \cite{lipman2023flow} trained on 5{,}000 and 150 demonstrations, respectively, from teleoperated and scripted data combined. Push moves objects up to 30\,cm and Pull moves a bin toward the robot. These skills are imperfect, succeeding on only 55.7\% and 39.7\% of random invocations, respectively.

\subsubsection{Learned Skill-Effect Model} The cheap model is an ensemble of three networks of hidden width 64, trained for 300 epochs with Adam (learning rate $3\times10^{-3}$, weight decay $10^{-5}$) on approximately 35{,}000 skill proposals logged from planner runs. All proposals train the success head with binary cross-entropy loss, and the roughly 20{,}000 physically valid outcomes train the effect heads with a smooth-$\ell_1$ loss. 

These data are collected offline for the fixed semi-structured environment, and the same model is used across all planning instances, amortizing the training cost across runs. Training and evaluation use the same workspace and object types, but evaluation environments feature independently randomized object subsets, poses, and clutter configurations from those seen in training. Scoring a batch of $K = 128$ skill proposals takes approximately 0.7\,ms on the GPU.

\subsubsection{Implementation Details} We use $K = 128$ proposals per oracle query, $N_{\mathrm{stall}} = 20$, random-certification probability $p_\mathrm{rand} = 0.01$, and fixed cost weights $(w_1,w_2,w_3) = (2,1,50\,\mathrm{m}^{-1})$ in Eq.~\eqref{eq:opt-cost} across all tasks and difficulty levels. The cost weights were fixed once based on the relative scales of the three terms, rather than tuned per task: $w_2=1$ retains the natural scale of the $-\log p$ penalty, while $w_3=50\, \mathrm{m}^{-1}$ makes 2 \,cm of predictive position uncertainty contribute one unit of cost; $w_1=2$ provides a comparable penalty for skill execution. Certification uses $\epsilon_p = 2.5$\,cm, $\epsilon_R = 0.087$\,rad and $\epsilon_q = 0.25$\,rad for object-position, object-orientation, and joint-space error, respectively. All experiments were run on a workstation with an Intel® Core™ i7-14700F CPU and NVIDIA GeForce RTX™ 5060 Ti GPU.

\subsubsection{Baseline Algorithms} 
We compare against \mosaic{} \cite{mosaic}, \emph{Options} \cite{konidaris2009skillchaining}, which chains skills forward from the start, \emph{CEM} \cite{cem}, which optimizes skill sequences in a receding horizon, and \emph{Roadmap}, a PRM-like skill-space planner \cite{PRM} without an oracle. We also give each baseline the same learned model. \emph{Lazy Options} and \emph{Lazy Roadmap} first search for a complete plan using learned skill-effect predictions. They then validate the plan by simulating its skills. If a skill fails, they remove the corresponding transition from the search graph and search for another plan. \emph{WM-CEM} predicts sequences and simulates the first step before replanning. 

Ranked-\mosaic{} scores the same $K = 128$ proposals but immediately simulates the highest-ranked proposal, isolating learned prioritization from lazy deferral; \lmosaic{} without uncertainty removes $u$ from Eq.~\eqref{eq:opt-cost}; WM-\mosaic{} executes the first complete predicted plan without certification.


\subsection{Results}
\label{sec:results}

Table~\ref{tab:planning_without_timeouts} reports the planning results. \lmosaic{} matches or exceeds \mosaic{}'s success rate in all eight settings, solving 104 of 160 trials compared to 81 for \mosaic{}. Restricted to successful trials, \mosaic{} averages 124, 331, and 479\,s on the easy, medium, and hard settings that both solve, against 43, 83, and 216\,s for \lmosaic{}, a per-setting speedup of up to 6.6 times (Task~3 Medium), with comparable solution lengths.

The other baselines solve fewer problems as clutter increases. Giving them the learned model reduces simulator queries (Table~\ref{tab:sim-skills}), but success rates remain low. Across the medium and hard settings, for example, Lazy Options and Lazy Roadmap obtain 9 and 13 successful trials, respectively, compared with 64 for \lmosaic{}. This suggests that the skill discovery strategy remains important even when skill evaluation is cheap. Task~3 remains particularly challenging: on its hard setting, \lmosaic{} solves 4/20 trials, while the no-uncertainty ablation solves 1/20 and all other methods solve none.

Ranked-\mosaic{} isolates the effect of learned proposal prioritization from lazy deferral. It reduces simulator queries from 46.4/67.9/134.5 for \mosaic{} to 44.3/62.9/107.9 on easy/medium/hard settings, whereas \lmosaic{} requires only 16.9/21.7/41.3. Learned prioritization therefore improves proposal quality, but the majority of the reduction in expensive evaluation is provided by lazy deferral. This deferral supports much larger optimistic graphs (Fig.~\ref{fig:opt-cert}): on hard settings, \lmosaic{} generates an average of 1{,}736 optimistic vertices and 587 edges while certifying only 17 vertices and 8 edges.
Prediction errors occur frequently enough that repair is regularly exercised: 14.0\% of successful certifications invalidate downstream proposals, triggering optimistic-graph repair. These repairs remain local, invalidating only 1.80 dependent proposals on average. Removing uncertainty also degrades planning performance, reducing total success from 104/160 to 90/160 while increasing simulator queries from 16.9/21.7/41.3 to 19.0/25.9/50.7 across difficulty levels. On held-out environments, the success head achieves PR-AUC 0.809 and Brier score 0.094, while mean and 90th-percentile object-position errors are 1.32 and 1.97 cm. Ensemble disagreement correlates with success-prediction error at 0.891 and position error at 0.614, supporting its use in the optimistic cost.

Removing certification makes WM-\mosaic{} fast on the trials it solves (1.7--8.1\,s), but only 33 of its 160 executed plans reach the goal, compared to 104 for \lmosaic{}. This shows that learned effects are useful for guiding search, but are insufficiently reliable to replace simulator-based certification.

%% file: sections/conclusion.tex
\section{Conclusion}
\label{sec:conclusion}

\lmosaic{} extends lazy search to skill-centric planning, where deferring a skill also leaves the state it produces unknown, so that vertices as well as edges are hypotheses. It searches over the outcomes predicted by a cheap learned skill-effect model, certifies the most promising skills in physics simulation, and, when a hypothesis is invalidated, repairs only the part of the graph that depended on it. Across eight cluttered-shelf settings, \lmosaic{} matches or exceeds \mosaic{}'s success rate, reduces simulator queries by 64--69\%, and plans 2.1--6.6$\times$ faster on successful runs in settings solved by both. Ranked-\mosaic{} shows that these gains do not arise from learned proposal ranking alone, while WM-\mosaic{} confirms the importance of simulator certification for reliable plans. In this work we target semi-structured settings, where the workspace and the object types are known in advance and the cheap model can be trained offline for them; extending \lmosaic{} to settings that require generalizing across workspaces is a natural next step. Within this scope, the accuracy of the cheap model bounds how much can be gained from laziness, and the cost of maintaining and repairing the optimistic graph grows with its size, accounting for part of the smaller speedup on the hardest settings. A more accurate model would ease both, and would let \lmosaic{} rely on certification less: planning and execution could be interleaved, with the robot acting on a certified prefix while the rest of the plan is still being searched, and, for tasks that are not safety-critical, dropping the expensive model altogether would leave a planner fast enough to replan continuously as the robot acts.